\documentclass[11pt]{article}
\usepackage{eacl2014}
\usepackage{times}
\usepackage{url}
\usepackage{latexsym}
\usepackage{booktabs} 
\usepackage[table,xcdraw]{xcolor}
\usepackage{csquotes}
\usepackage{mathtools}
\usepackage[british]{babel}

\usepackage{etoolbox}
\makeatletter
\apptocmd{\@chapter}{\expandafter\label\expandafter{\unexpanded{#2}}}{}{}
\apptocmd{\@sect}{\expandafter\label\expandafter{\unexpanded{#8}}}{}{}
\makeatother

\usepackage{hyperref}

\makeatletter
\DeclareRobustCommand*\textsubscript[1]{%
  \@textsubscript{\selectfont#1}}
\def\@textsubscript#1{%
  {\m@th\ensuremath{_{\mbox{\fontsize\sf@size\z@#1}}}}}
\makeatother

\title{Notes About a More Aware Dependency Parser}

\author{Matteo Grella \\
  Independent scholar \\
  {\tt matteogrella@gmail.com} \\}

\date{}

\begin{document}
\maketitle
\begin{abstract}
In this paper I explain the reasons that led me to research and conceive a novel technology for dependency parsing, mixing together the strengths of data-driven transition-based and constraint-based approaches. In particular I highlight the problem to infer the reliability of the results of a data-driven transition-based parser, which is extremely important for high-level processes that expect to use correct parsing results. I then briefly introduce a number of notes about a new parser model I'm working on, capable to proceed with the analysis in a ``more aware'' way, with a more ``robust'' concept of robustness.
\end{abstract}

\section{Introduction}

To ease the reading of this article I have decided to adopt a general descriptive approach as much as possible to present my research as well as its underlying motivations, rather than a more specific and formal one that will be used in next papers.

The structure of this paper is as follows: I start with background information on Natural Language in section ~\ref{Natural Language} and then on Dependency Parsing in section ~\ref{Dependency Parsing}. In sections ~\ref{Data-driven transition-based approach} and ~\ref{Constraint-based approach} I'll focus on two approaches that fulfil the dependency parsing task, respectively data-driven transition-based approach and constraint-based approach. Section ~\ref{Motivations for a new approach} explains the motivations for a new approach. The final section ~\ref{Notes about a new approach} briefly introduces a new hybrid parser model I'm working on that proceeds with the analysis in a ``more aware'' way, whose meaning is explained in the next sections.

\section{Natural Language}\label{Natural Language}

Natural Language is a very complex system, involving many brain processes. Trying to reproduce it by means of artificial agents has been one of the main goals of Artificial Intelligence since its early days. For more than 50 years, linguists and computer scientists have tried to make computers understand human language fighting against its fascinating and misleading nature: implicit, highly contextual, ambiguous, often imprecise and contingent to biological processes. Indeed the language appears subject to the whims of evolution and cultural change on one hand, and based on strong rules that constrain the possible sequences of phonemes or words on the other\footnote{For example, in written language the sequence Article + Article (the the cat) is meaningless if not unacceptable.}.

As a matter of fact it is hard to deny that linguistic production and comprehension are based on a system of formal regularities, and that some of these regularities have a stable behaviour in a given moment of the historical evolution of a given language. Furthermore, these regularities are precisely what legitimate the use of the term ``system'' when one talks about language. These ideas are well explained by a quote from De Mauro (1990) whose a free translation follows: ``On one hand there are, well founded, the reasons that assimilate the language to a calculation. On the other hand, no less strong, there are the reasons that preclude such assimilation. Theorists and philosophers generally have opted for accentuation of one or the other. (...) It seems to us that a good theory of the language must take into account those and the other reasons''.

This calculation vs. non-calculation dilemma is the reason why - to name but one - Chomsky (1965) distinguishes between competence system and performance system, giving rise to principles such as grammaticality and acceptability of a sentence as well as the distinction of its surface and deep structure. While these issues attract a lot of attention from a theoretical point of view, they are not so relevant for the realisation of a practical system for natural language analysis: what is important is to acknowledge that gradation is a central phenomenon in natural language, which means that not all sentences fit in to the binary distinction of grammatical vs. ungrammatical; some are simply ``slightly better'' than others without the latter having to be completely rejected.

From a computational perspective I can really say that this dichotomy was the root of the ``theory split'' that still today characterizes most of the studies and the research on Natural Language Processing (NLP), split as we said into two branches that for simplicity I can call \textit{rules} and \textit{statistical} based approaches.\footnote{Here the term \textit{statistical} is improperly used because we also include probabilistic approaches.}

\section{Dependency Parsing}\label{Dependency Parsing}

Dependency Parsing is an attractive alternative to constituency parsing for syntactic analysis, commonly considered one of the fundamental steps for linguistic processing because of its key importance in mediating between linguistic expression and meaning.

The theory behind dependency parsing is based on the Dependency Grammar which can be proud of a long-standing tradition in linguistics: several theories and formalisms (Tesni\`ere (1959), Sgall (1986), Mel'\v{c}uk (1988), Hudson (1990), Maruyama (1990)) share the fundamental assumption that syntactic structure consists of word-to-word dependencies i.e. lexical nodes linked by binary asymmetrical relations called dependencies. Dependency Grammar is sometimes called Valency Grammar, a name conceived by analogy with chemical valency, according to which some words (especially verbs) have valencies dependent on the number of elements (e.g. nouns) with which they combine.\footnote{Because of this analogy, sometime it is possible call lexical nodes ``atoms''.}

In a dependency structure, every word is dependent on, at most, another word (its governor).\footnote{Alternative terms in the literature are \textit{regent} and \textit{head} for \textit{governor}, and \textit{modifier} or \textit{argument} for \textit{dependent}.} This means that the structure can be represented as a dependency tree, where nodes are words and arcs are dependency relations (e.g. subject, direct object, modifier). Another requirement for a well-formed dependency tree is that there is precisely one root, which is usually the main verb of the sentence (as a consequence of the ``verbo-centricity'' theory). Thus, the task of a dependency parser is to take a sentence (input text) represented by a sequence of words (nodes) and enrich it with the appropriate set of labeled dependency arcs. Each labeled dependency arc involves exactly two words and a label.\footnote{The only exception is for the root node which can have a label but not a true governor.} More formally, dependencies can be represented as a set of directed arcs of the form \textit{g} $\xrightarrow{\textit{l}}$ \textit{d}, where \textit{g} is the governor node, \textit{d} is the dependent node (\textit{g} $\neq$ \textit{d}) and \textit{l} is the label, resulting in a dependency structure called dependency tree (parse tree). For more details on dependency tree, dependency grammar and dependency parsing see Nivre (2003) and the references cited therein.

As we will see later, this is an over-simplification, if nothing else because not all the words that should be considered are expressed in the input text: there are ``words'' hidden in the surface structure but essential to keep up a syntactic structure. This happens because understanding a sentence means to translate the linear order mainly originated for physical reasons (during which some elements may be lost) into a structural order, bringing back its original hierarchical structure (Tesnière, 1959).\footnote{Time-linearity in spoken language and space-linearity in written language.} I'd like to think that if we could communicate ``telepathically'', without physical constraints (linear sequence of words), then the information would be transferred directly from one brain to the other as a structure similar to a dependency tree, with all the elements naturally hierarchically organized.

Back to the main subject, several approaches have been developed to fulfil the dependency parsing task. In this paper I'll focus on the data-driven transition-based approach and on one of its ``counterpart'', wcdg-parsing, based on the Weighted constraint dependency grammar (WCDG), which is grounded on a specific descendant of the constraint-based approach (Heinecke et al. 1998; Schr{\"o}der, 2002). These two approach are representative of both \textit{statistical} and \textit{rules} worlds respectively, and both of them achieve similar overall state-of-the-art results. See Krivanek and Meurers (2011) for a comparison of a statistical and a rule-based dependency parser.

An overview of the two approach will be given in the following sections, introducing crucial key concepts to understand the motivations for a new parser and the technology that could be adopted for its implementation. 

\section{Data-driven transition-based approach}\label{Data-driven transition-based approach}

I assume the reader is familiar with the formal framework of transition-based dependency parsing originally introduced by Nivre (2003).

To summarize, transition-based parsing is based on a transition system that processes the input sentence by means of transitions which incrementally build the dependency tree. The sequence of transitions is called computation. The system is initialized to an initial configuration based on the input sentence, to which transitions are applied repeatedly generating new configurations until the final configuration is reached. Given a configuration, transitions can create a dependency arc between governor and dependent nodes. Transitions that create arcs \textit{encode} in itself several information, including arc direction (left vs. right), from which it is possible to identify the involved nodes, and a label representing the name of the syntactic relation.

Data-driven means that there is no need of a hand-written grammar and the analysis process is guided by the words (the data) of the sentence itself. This approach takes advantage of the increased availability of dependency treebanks (i.e. sentences manually annotated in parse tree format) and of the recent techniques to apply machine learning algorithms to natural language processing to implement a training procedure able to generate an ``inductive grammar".

During the learning phase, the treebank sentences are processed and the parser learns how to use the transitions emulating the transitions sequence implemented in the \textit{oracle} function by means of a classifier, so that learning a grammar means learning to select what is the best next transition giving a configuration. \textit{Oracle} is the name given to the function that maps a parser configurations to optimal transitions with respect to an annotated sentence (gold tree). Giving a configuration /transition pair, a set of features is extracted from the configuration \textit{summarizing} it and used to properly train the classifier.

\textit{Features} are one of the most important type of information that data-driven systems use to lead the analysis process. After the training phase (whereas the transitions are established in advance by the oracle), these features must be sufficiently robust to guide the parsing process without a complete view of the sentence. Data-driven transition-based parsers use only contextual information i.e. a limited window of nodes, including the already parsed ones, centred around the focus point of the analysis. Features that take into account these elements can encode combinations of several words properties e.g. forms, pos-tags, arc labels of the words itself or of its left and right dependents. Zhang and Nivre (2011) proposed a rich features template considering third-order features, linear distance between a pair of possible governor and dependent, valency informations. 

This approach is represented by the models of Yamada and Matsumoto (2003), Nivre (2003, 2004), Attardi (2006), Nivre (2009), Goldberg and Nivre (2013), Sartorio (2013).\footnote{The arc-standard and arc-eager models are two of the most widely known and used transition-based system.} These models mainly differ by the way they define configurations i.e. set of available transitions as well as the ability to handle discontinuous syntactic constructions.\footnote{Discontinuous syntactic constructions are also called non-projective dependency structures because of the presence of crossing edges.} 

As a side effect of their incremental behaviour, data-driven transitions parsers have limited look ahead capabilities (i.e. they are limited to local features) so they are affected by the problem of having to decide sometimes too early how to proceed with the analysis, before having seen the remaining part of the sentence, so they likely make mistakes, especially on long distance dependencies (Bohnet, 2011). This search errors cause error propagation (McDonald and Nivre, 2007) i.e. when the parser makes an error, the probability that it makes others increase because it enters into configurations for which it has not been trained so it does not know how to react (Goldberg, 2013). The most common approach is to use beam algorithm instead of exploring only a single derivation for each input (greedy decoding). The drawback of the use of the beam search is that parsing speed are not fast as the original greedy transition-based parsers. 

Recently, several changes have occurred with respect to the original approach in order to mitigate this error propagation problem without altering parsing time. The most prominent approach is to use \textit{dynamic oracles} in combination with online-learning techniques, enabling an \textit{error-exploration} procedure that improves the way in which classifiers learn from data. In essence, error-exploration consists in exposing the training procedure (then the classifier) to non-optimal configurations (computations that do not lead to a gold tree) obtained following sometimes erroneous predicted transitions together with the \textit{optimal transitions} for those configurations, providing the parser with a sort of self-consciousness of its own mistakes. Parsers that exploit these flexible oracles achieve state-of-the-art results for greedy parsing, with a big difference in terms of accuracy compared to static oracles -typically 1-2$\%$, with no differences in parsing time, obtaining scores comparable to those of the best statistical graph-based parsers (McDonald et al., 2005).

These parsers usually work monotonically since arcs are only added to but never removed from the set of dependencies. Honnibal (2013) suggests an ``error-repairing'' strategy, implemented as a non-monotonic version of the arc-eager system, which combines error-exploring technique with some relaxing of the transitions preconditions, allowing the parser to recover the correct arc from the wrong governor assignments forced by the past incorrect transitions. Although the idea of error-repairing is very interesting, its real recovery capabilities are very limited as for now, providing an improvement of up to 0.2$\%$ accuracy (Honnibal, 2013).

\section{Constraint-based approach}\label{Constraint-based approach}

In this section I'll focus on wcdg-parser (Foth and Menzel, 2006), a mature implementation of wcdg-parsing based on the WCDG grammar. WCDG extends the CDG formalism first described by Maruyama (1990), and it was demonstrated to be appropriate for modelling a large variety of linguistic phenomena such as immediate dominance, agreement, valence, aspects of word order and projectivity.

In this approach the parsing problem can be viewed as a constraint satisfaction problem (CSP). A recent introductions to CSP can be found in Miguel and Shen (2001). The dependency parsing main process is seen as the problem of finding a dependency tree for a sentence that satisfies the constraints defined by a hand-written grammar.

The WCDG main aspect concerns the possibility to express graded constraints rather than hard grammar rules: to each constraint is assigned a weight or \textit{penalty} between 0.0 and 1.0 that indicates its importance. The weight 0.0 is associated to hard constraints which theoretically can only be violated when no other solution is possible, while different weights (soft constraints) assign preferences among many linguistic phenomena. Furthermore, the formalism of WCDG provides dynamic constraints which do not have a static score but receive different weights depending on the con text in which they are evaluated. Usually the constraints and their related weights are determined by the grammar writer. Recent works attempted to compute the weights of a WCDG automatically by observing which weight vectors perform best on a given corpus (Schroder et al., 2001), but weights computed completely automatically failed to improve on the original hand-coded grammar.

For instance, constraints can express that:

\begin{itemize}

\item preferably the top node is a \textit{verb} (soft constraint);

\item preferably the top node is a \textit{finite verb} (soft);

\item a node does not have more than one \textit{object} (hard constraint);\footnote{A special dependency label, called \textit{extra-obj}, could be used in order to allow \textit{ripresa pronominale} due to \textit{dislocazione a sinistra}, terms borrowed from the Italian language where is very frequent this type of syntactic constructions.}

\item \textit{determiners} must precede their governor (hard) and that it is most often a \text{noun} (soft); or that there cannot be two determiners for the same governor (hard); or that a \textit{determiner} and its governor must agree in \textit{number} and {gender} (soft). 

\item an \textit{article} modifies a nearby \textit{noun} (dynamic constraint);

\end{itemize}

Wcdg-parser uses beside constraints an information-rich (e.g. valence) hand-crafted lexicon (Foth, 2006). Further details can be found in Foth (2004) and Foth and Menzel (2006).

Since the general CSP is an NP-complete problem, also wcdg-parsing can result in non-termination and efficiency problems.\footnote{Indeed, some solutions of language processing algorithms that would be ideal in theory have a complexity that corresponds to the NP-Complete problems: a trade-off exists among the solutions theoretically most elegant and the solutions that can be implemented practically.} Instead of a full search, wcdg-parser uses a heuristic search called Frobbing, a non-monotonic transformation-based constraint resolution method with anytime properties (Foth et al., 2000).\footnote{Anytime property: the parser maintains a complete analysis at any time so the algorithm it can be stopped at any time and return a complete analysis. Anyway there is a trade-off between parsing time and quality of results so the time left for an analysis generally coincides with a better accuracy.}

Wcdg-parser tries to find an analysis (a dependency tree) by transforming a given one until it cannot be improved further. In Frobbing, an arbitrary dependency structure is constructed first from the input sentence, then the algorithm tries to correct analysis errors selecting \textit{transformations}, based on constraints that cause conflicts (constraints that are violated for specific dependency arcs). Given an analysis, transformations generate new analysis changing a set of local properties such as a label or governor of a dependency relation as well as a pos tag of a node or its morphological feature (e.g. case, gender, number, mood, tense, etc.). A set of conflicts is then recomputed and the most severe (weight close to 0.0) of them is \textit{attacked} by transforming the analysis. This results in an analysis which is not necessarily better (i.e. with less severe conflicts, since other conflicts may be created in this step) but that does not have that specific attacked conflict anymore. If the conflict can not be removed, the algorithm tracks back to the last starting analysis - resulting in a search strategy similar to tabu search. The whole process is repeated until a new better analysis is found and marked as the new starting analysis. The algorithm ends when no other analysis improvements are possible.

The constraint-based approach is useful especially for richly inflected languages and free word order such as, for example, Italian and German which, according to recent experiments, have a syntax considerably more difficult to analyse than English. Dubey and Keller (2003) and Grella (2011) show better results for constraint-based parsers with respect to statistical parsers, respectively, in an evaluation on NEGRA treebank (Brants et al., 1999) for German and TUT Treebank for Italian (Bosco et al., 2000) . 

\section{Motivations for a new approach}\label{Motivations for a new approach}

Nowadays, and even more in the future, an effective syntactic parser is involved in any cutting-edge application devoted to text processing (document indexing, information extraction, automatic translation, sentiment analysis, ...), so the parser must be \textit{robust} i.e. able to produce analysis for any type of input including \textit{non-canonical} material such as spoken language transcription or e-mails which may include syntactically ill-formed sentences. Another crucial feature of syntactic parsers is the efficiency due to the increased availability of data (the \textit{Big Data}) to analyse.

The syntactic parser output is the starting point of other high-level processes which expect to use correct parsing results, so it is extremely important to be able to predict the reliability of the results of a parser. Otherwise, using incorrect parsing information, a degradation of the applications performance is almost guaranteed.

Among the advantages of data-driven transition-based parsers there are state-of-the-art accuracy and the linear time complexity of many of them. Greedy parsers are the fastest approach for dependency parsing, enabling web-scale parsing with high throughput.

This parsing approach seems appealing not only from an engineering perspective due to its efficiency, but also from a psycholinguistic point of view as they process a sentence incrementally much the way that people do, thing that has motivated several studies concerning their cognitive plausibility (Nivre, 2004; Boston and Hale, 2007; Boston et al., 2008). 

From a cognitive prospective, the data-driven statistics nature puts this approach inline with Bod (2003):

\blockquote{Language displays all the hallmarks of a probabilistic system. Grammaticality judgments and linguistic universals are probabilistic and stochastic grammars enhance learning. All evidence points to a probabilistic language faculty.}

and with Norvig (2011):

\blockquote{It seems clear that probabilistic models are better for judging the likelihood of a sentence, or its degree of sensibility. But even if you are not interested in these factors and are only interested in the grammaticality of sentences, it still seems that probabilistic models do a better job at describing the linguistic facts.}

Unfortunately, it is difficult to infer the reliability of the results of a data-driven transition-based parser and this assumption does not fit well in case of actual implementation of statistical dependency parsing: there is no straightforward mapping between the parser output score, if any, and some simple notion of grammaticality.\footnote{Despite this, in the context of Domain Adaptation with Active Learning, Attardi (2011) uses the score that the parser itself provides as a useful measure of the \textit{perplexity} in parsing a sentence.} On the contrary, Fong and Berwick (2009) found that, despite their results, such parsers fail to incorporate much ``knowledge of language'' in many cases: they fail to replicate many empirically attested grammaticality judgments; seem overly sensitive, rather than robust, to train data idiosyncrasies; and easily acquire ``unnatural'' syntactic constructions.

A possible explanation is that usually the data-driven dependency parsers drop accuracy in domains outside the data from which they were trained, and the enthusiasm generated by the ``underlying semantics'' that seems to be assimilated in this model is revealed actually quite fragile.\footnote{How to increase the accuracy of a parsing system when dealing with out-of-domain texts is the goal of \textit{domain adaptation} task. Usually, techniques such as self-learning or active-learning (Attardi, 2013) are used.} The reasons for this can be found in some different distributions of morpho-syntactic features extracted from the set of sentences in the treebank which, despite they are numerous, are anyway limited in number and typology with respect to the \textit{general language}. For example, the famous Penn Treebank corpus (Marcus et al., 1993) is one of the largest treebank but it is dominated by financial news from the Wall Street Journal that contains quite a peculiar linguistic phenomenon as journalistic expressions. More trivially, state-of-the-art dependency parsers use a highly sparse lexicalized model: it means that the features are created using word forms and lemmas (when available) so that co-occurrences of certain words in the given treebank are combined into lexicalized syntactic \textit{ngrams} features (dependency tree fragments). Therefore, ideally all possible valid word combinations that the parser will face during parsing should be recorded in a treebank, which is unlikely to happen if we consider both the limited size of this resource and the fundamental onniformativity principle of natural language, according to which languages may express any learning experience (De Mauro, 1990). On the other hand, in the treebanks there are syntactic and distributional homonymous structures (John Lyons, 1970); in other words the same surface structures (i.e. pos sequences) can match very different syntactic analysis, depending on the basis of semantic relationships among words. For this reason, homonymous structures cause troubles to a delexicalized parser, with an accuracy difference between lexicalized and not lexicalized parser greater than 6$\%$ in the test set of the same domain of the training set, showing that lexicalized features seem indispensable. Basically, lexical information are important but too sparse. Techniques have recently been introduced to broaden the spectrum of words and not to limit the feature identity to a match of an exact word. Koo and Carreras (2008) improved parsing accuracy and coverage substituting the words form with an attempt to ``semantic words knowledge`` in the form of clusters which merge words according to contextual similarity extracted from very large corpus.\footnote{In binary representation, clusters serve as coarse lexical intermediaries and are equivalent to bit-string prefixes from which prefix length determines the granularity of the clustering e.g. 01 fruit, 010 apple, 011 orange (e.g. Brown's cluster algorithm (1992).} ``Cluster-based'' feature sets have progressively boosted by recent distributed word representations (word embeddings) where each word is represented by a dimensional dense vector, instead of clusters encoded in static strings (Collobert et al., 2011). As for clusters, word embeddings are learnt from large corpus in such a way that concepts with similar or related meanings are near each other in that space i.e. similar words are expect to have close vectors.\footnote{This is possible because word embeddings can be induced directly from widely available unannotated corpora of different domains otherwise not covered by traditional linguistic resources.} Chen and Manning (2014) developed a transition-based parser using a deep learning architecture which exploits word embedding as features and also creates a dense vector representation for pos tag and arc label instead of a discrete representation. Their parser, as well as the Attardi's (2009) and Grella's (at Evalita 2014) ones, take advantage of the neural network architecture (multilayer perceptron) that already incorporates non-linearity in the hidden layer to infer the interaction starting from ``atomic features'' (one features for each element properties), so that a manually designed feature template (e.g. pairs or triplets of word properties) that provides a form of non-linearity useful for linear classifier like averaged perceptron, is no longer required.\footnote{A manually designed feature template, by definition, suffers from the following problems: sparsity, incompleteness, expensive features computation.} 

These techniques performed well with an averaged of +0.7 with respect to traditional lexicalized model, nevertheless a strong dependence remains with the context in which the parser has been trained, and they don't solve at all the problem of distinguishing, as people do, between acceptable and not acceptable sentences.

Furthermore, in the case of \textit{ungrammatical} input sentence there is no guarantee that data-driven transition-based parsers yield the ``right'' wrong analysis as the most probable analysis. If we consider that also with a correct \textit{grammatical} input it is not possible to determine the reliable level of confidence of the analysis in output, to distinguish among correct and incorrect parser results seems to be not feasible.

All this makes it quite evident that the presence of a grammar (rules or constraints) is essential. In this direction, as mentioned in a previous section, a competitive wcdg-parser has been developed for unrestricted German input that is largely independent from domain and achieves state-of-the-art results. The wcdg-parser, thanks to a grammar that ``extended'' the notion of grammaticality, is able to produce an effective score that can be used to determine the degree of acceptability of a given analysis, together with an accurate characterisation of the input text by means of a list of unremovable soft and hard violated constraints.

Constraint scores also help to guide the parser towards the optimal solution and allow the parser to deal with the input in a robust way. Anyway, since Frobbing is a heuristic procedure, at the end of the algorithm there's no certainty that the optimal solution has been found. This means that it sometimes fails to find the correct dependency structure of an input sentence even if the language model (i.e. the entire set of constraints including lexical information) accurately defines it, because of search errors during heuristic optimization. In addition to this, although many defeasible constraint exist which allow but disprefer certain construction, there are many more possible but implausible dependency structure that are not dispreferred. A reason for this is that sometimes some possible syntactic constructions are distant from any structural ambiguity perceptible by a human and so it is difficult to conceive them before they are computed by the parser. This is part of a typical issue of any hand-written grammar formalism: to write rules by hand about every \textit{noun} or \textit{verb} of a language, which seems more and more necessary once one gets closer to more specific language phenomena, is simply infeasible.

In any case, efficiency of wcdg-parsing as it now is, is clearly not competitive with that of statistical dependency parsing, with parse times of several minutes, or, even worse, of hours in case of some \textit{complex} sentence, making this approach unusable in actual context.

The matter has now arrived to the point of Norvig and Chomsky debate (2011), that recall the natural language properties described in section ~\ref{Natural Language}. Norvig suggests that ``probabilistic, trained models are a better model of human language performance than are categorical, untrained models'', meanwhile Chomsky objects that ``It's true there's been a lot of work on trying to apply statistical models to various linguistic problems. I think there have been some successes, but a lot of failures. There is a notion of success ... which I think is novel in the history of science. It interprets success as approximating unanalyzed data. (...) We cannot seriously propose that a child learns the values of 109 parameters in a childhood lasting only 108 seconds.".

With a clear understanding of the limitations and benefits of both data-driven and constraint-based approach, in the next section I briefly introduce a technique I'm working on to build an innovative fast hybrid technology for dependency parsing that combines the strength of both of these approaches, with the objective to create a new parser that is able to proceed with the analysis in a ``more aware'' way.

Before leaving this section where I tried to explain the reasons for a new approach, it is important to notice that, regardless the approach used, architectures derived from the research in NLP traditionally separates the process of language analysis into a series of more simple tasks executed in sequence, sacrificing the advantages of possible parallelism, as in many situations it would be beneficial to exploit information being produced by one task while performing another task. Moreover, this architecture may suffer from error propagation problems, especially when clearly interdependent tasks are modelled separately, as in the case of the lemmatization, part-of-speech tagging and syntactic parsing. For instance, a typical model of syntactic parser presupposes that input words have been morphologically disambiguated using first a lemmatizer, then a part-of-speech tagger before parsing begins. This is bad especially for richly inflected languages such as - among others - French, German, Italian and Spanish (also known as morphologically rich languages), where there is a considerable interaction between morphology and syntax such that neither can be fully disambiguated without considering the other.

Even the boundaries between what is the domain of syntax and what is the domain of semantics is very thin: one can just take a look at the well-known PP-attachment problem or the correct identification of several conjuncts involved in a coordination chain, the anaphoric references and more.\footnote{More generally, a well known problem in parsing tasks is the dependency ambiguity: for a given sequence \textit{A-B-C}, both interpretations \textit{A(B(C))} and \textit{A((B)(C))} may be structurally possible.} Some (yet fundamental) words can be omitted in the surface structure when they are to some extent implicit or semantically inferable (e.g. because of null elements resulting by pro-drop). All this suggests that the syntactic and semantic analysis should be performed together. To have a resulting parse tree \textit{complete}, the mechanisms that guarantee this completeness must find their place and role in the parsing process. These mechanisms should include traces integration and words sense disambiguation.

\section{Notes about a new approach}\label{Notes about a new approach}

I think that, while a probabilistic component is essential to resolve ambiguity in both syntax and semantics, it is also crucial to equip the analysis system with a \textit{stable} linguistic knowledge. My premise is that in the natural language it seems it is possible to distinguish between possibility and probability of a given utterance.

As a matter of fact, the idea of combining linguistic rules based component with a statistical engine is widely used, especially for machine translation systems. In a dependency parsing context, a method have been proposed to add statistical components as ``oracles'' to a constraint-based parser: Foth and Menzel (2006) developed a hybrid version of the wcdg-parser, which uses a probabilistic transition-based parser as an initial statistical predictor component. In their model, the output of the statistical parser (Nivre 2003) is converted to soft constraints which encourage the constraint solver to create first the same dependencies of the statistical parser, leaving then to the Frobbing algorithm the hard work to find a better solution only if the dependencies created by the first system generate conflicts.\footnote{A weight of 0.9 has proved to work best on an evaluation using the sentences 501 to 1000 of the NEGRA corpus.} They proved that the use of a statistical parser enables the wcdg-parser to produce a better initial attachment (which means that less time has to be used to correct attachment errors), and that such mixed approach is useful to reduce some characteristic issues such as modeling and search errors, in particular for long and complex sentences. The reason for this is that correct \textit{easy} attachments are rather common, and those that require deeper analysis are comparatively rare.\footnote{Here the concept of \textit{easy attachment} is borrowed from Goldberg and Elhadad (2010) i.e. the arcs that the statistical predictor can get with high reliability, as \textit{noun} $\xrightarrow{\textit{det}}$ \textit{article}.} However, the system remains slow, with a sentence analysis times of the order of seconds.

I suggest a different approach from Foth and Menzel (2006): in a new parser, the two components could be used at the same time and not sequentially as a pre- or post-processing of the other component. My aim is to incorporate linguistic insights (weighted constraints) into a fast data-driven transition-based system: the idea is that the constraints mainly control the possibility of syntactic expressions (grade of acceptability respect to the grammar) and the statistical component uses their \textit{probabilities} (i.e. the grade of confidence which encode a sort of language competence according to the statistical model trained on corpora) to dynamically guide the parsing process to the solution. At the beginning I was fascinated by the human behaviour of a data-driven transition-based parsers. Today I think that such parsers are more useful to navigate the search space in order to avoid the evaluation of unlikely solutions, than heuristic searches such as Frobbing can do. 

I think also that a non-monotonic behavior, obtained through back-tracking, transformations or beam search, is fundamental to combine in the same process more tasks such as pos-tagging and dependency parsing, exploring a larger space that best suits to the amount of ambiguity to be considered. In order to mitigate the error-propagation problem, the new parser could compute a number of alternative syntactic structures in parallel using a beam search algorithm (parallel parsers) sharing information among different analysis, in contrast with the parsers that compute only a single preferred analysis (serial parsers). For efficiency reason, the beam size may be limited to 10 so the parser would explore a very small fraction of the many possible analyses whose number grows exponentially. Actually, I'd like to think that in human reading process few alternatives are retained until the ambiguities and uncertainties are resolved. And this, beyond the obvious reasons of efficiency, seems to me more reasonable than a beam size of 64 used by Zhang and Nivre (2011).

Despite the use of a beam search, no proper analysis could be found in case of difficult or ill-formed constructs that always violate a hard constraint. To handle this, in the new approach I introduced the concept of \textit{unknown} arc, a dependency label used if the best syntactic connections found continues to violate a hard constraint and therefore there is no theoretical and formal justification to specify other label. A next level to syntactic parsing can then take into account this insight.\footnote{This is different from \textit{dep} stanford dependency that is used when the system (or a human) is unable to determine a more precise dependency relation between two words.}

Among the advantages, the new hybrid approach allows the use of constraints with a high level of abstractions. A subset of known universal linguistic knowledge is successfully used in unsupervised dependency parsing achieving state-of-the-art accuracy in that context (Naseem, 2010). This knowledge may be converted in constraints that could be used to control the execution of a transition given by the classifier.\footnote{Some parameters used by universal constraints (e.g. dominant subject, verb, object sequence order) should be learned during a first reading of the treebank.} By means of general universal constraints become easy to filter some implausible construction proposed by the statistical component e.g. the maximum degree of multiple center-embedding of clauses is exactly 3 in written language (Karlsson, 2007).\footnote{No real examples of degree 4 have been recorded. In spoken language, multiple center-embeddings even of degree 2 are so rare as to be practically non-existing (Karlsson, 2007).}

Practically, in my experiments the data-driven transition-based component is based on the arc-standard model (Nivre, 2004), extended with:

\begin{description}

\item[ a.] a \textit{$wait$} transition (Yamada and Matsumoto, 2003) used to create hypothesis (that take part of features) about some ``delayed dependencies'' during a shift, due to a pure bottom-up arc-standard strategy, approximating the behavior of the arc-right transition in the arc-eager model (Nivre, 2003) which create arcs in a more incremental way, in line with psycholinguistical point of view that postulate humans tend to make predictions about syntactic structure and process local attachments first (Gibson, 2000);

\item[ b.] the non-adjacency transitions proposed by Attardi (2006) to handle non-projective dependency structure maintaining linear time complexity. Even if the non-adjacency transitions have an incomplete coverage of non-projective structures, Attardi (2006) notes that a distance lower than 3 is sufficient to handle almost all cases of non-projectivity in the training data of almost all languages. This model takes advantage of a great intuition of Attardi (2014) to extend the use of non-adjacent arc transitions used so far only to handle non-projectivity, also for recovering a overlooked proper arc;

\item[ c.] a specific transition that considers the top node from the second to the last configuration, following the \textit{None} approach discussed by Ballesteros and Nivre (2012). In their model, if no dummy root node is added (at the beginning or at the end of a sentence), then there is no explicit transition to link the top node, meaning that the last node remaining in a configuration will be treated as root dependent. By contrast, in this model a specific \textit{root} transition is used to assign the top label to the top node. As a matter of fact only when the parser sees the whole tree, it can verify the ``integrity'' of the solution;

\item[ d.] a special mechanism integrated with arc transitions to treat punctuations, in a way similar to Ma et al. (2013). Traditional transition-based parser considers punctuations as well as words although these are not as consistently annotated in treebanks as words, making it harder to parse. In the new parser, before the creation of the initial configuration, some punctuations (e.g. commas) are first attached as a properties of their right neighbouring words, then removed from the input sentence. Such information are propagated from dependent to governor step by step and used as features; 

\item[ e.] an enrichment of the information encoded into transitions, in a way similar to Bohnet and Nivre (2012) who introduce a transition-based system that jointly performs pos tagging and dependency parsing encoding pos-tag information in the \textit{shift transition}. The new technique I'm proposing here consists in enriching the arc transitions (e.g. arc-left, arc-right) adding the pos-tag to the syntactic label (deprel), so the parser can easily performs with a single transition pos-tagging and syntactic analysis. Using this technique where dependency labels natively consist in pos-deprel combination (e.g. noun-subj, adj-rmod), words may have more than one deprel in order to handle ``agglutinate words'' without a preprocessing task;\footnote{For example in the Italian language, the word \textit{della (di+la)} may have a deprel \textit{prep-conn art-det}, and word \textit{leccala (leccare+la)} may have a deprel \textit{verb-top pron-dobj}.}

\end{description}

Technical details, necessary to explain these extension as well as the successive notes, will be the subject of a next paper.

The application of constraints occurs whenever the statistical transitions system proposes an arc creation. 

As a matter of fact, that is exactly the moment where traditional transition-based dependency parsers already impose certain constraints: not all transitions predicted by the machine learning algorithm are valid at each configuration, due to preconditions in the transition system, so before the execution only the valid transitions are sorted and the best one executed. Nivre, Goldberg and McDonald (2014) extend the transitions of the arc-eager model with preconditions for different constraints, in order to block some of these according to some fixed criteria. In their \textit{empirical case studies} they consider the problem of parsing commands to personal assistants such as Siri or Google Now. In this specific context it is plausible that if the first word of a command is a verb, it is likely the root of the sentence. Using a simple constraint as ``the first word of the sentence must be the root'' they have achieved an important accuracy improvement (over 3\%), demonstrating the effectiveness of the use of additional information sources not directly inferable at the training time. \footnote{Grella et al. (2011) during the workshop of Evalita 2011, in \url{http://www.evalita.it/sites/evalita.fbk.eu/files/presentations2011/Grella.pdf} present a technique similar to \textit{arc constraint} of Nivre et al (2014), called \textit{multilayer linguistic supervision} which use subcategorization information (e.g. transitivity) in order to block meaningless transition.}

The new parser uses the pos-tag information encoded in the transition as a precondition, in a way similar to Nivre et al (2014). For each arc transition, it is able to check if the set of all the possible available lexical readings of the target, the dependent node, contains the desired pos-tag.\footnote{The parser has to include a morphological analyzer, for example using a dictionary of word forms and multi-word-expression, with associated PoS, lemmas and grammatical features (e.g. mood, tense, person, gender, number, case).} If so only compatible lexical readings are maintained (the parser starts with all possible lexical readings for each word), otherwise the transition is rejected. This mechanism (which is an essential part of the online morphological disambiguation process) ensures that the dependent is morphologically valid, but nothing can be inferred about the validity of the governor, with respect to its new dependent, and then to the \textit{consistency} of the dependency relation just created.\footnote{This is more complicated in practice, since, starting from governor and dependent, the disambiguation process (i.e. the elimination of non-valid lexical readings) could involve nodes at an arbitrary depth.} However, the more a governor collects its dependents, the more the chances increase that a constraint could help to disambiguate also the governor lexical readings. For instance, the constraint ``\textit{verb-aux} depends only on a \textit{verb}'', restricts the possible readings of the governor, even constraining the realization of the dependency relation itself or reducing its reliability score.

In this model the constraints can act on several levels of analysis and they are not only able to filter (passive checks) but also capable to generate new information during the analysis process, which in turn may be used by the statistical component.

In addition to the expected boolean result of the application of a constraint, also a \textit{features transportation} can be obtained.\footnote{Boolean result means that a constraint is passed or violated. In the latter case it returns also the value of \textit{penalty} associated.} Feature transportation refers to the process of propagating some kind of information along the dependency tree. This mechanism is required to describe natural language phenomena where constraining information is applied at a particular node but originates from a structurally distant one. Since the distance may be of arbitrary length, the useful information are hardly contained in the set of local features which are visible from the statistical component, so it is important for a node to \textit{inherit} properties from another node. Features transportation could be view as a sort of ``unification'' mechanism widely used on unification-based grammar.\footnote{See Shieber (2003) for an introduction to unification-based approaches to grammar.}

Among others, in the new parser this mechanism is used in case of:

\begin{itemize}
\item determiners: a determiner may impose a gender, a number and a grammatical case to its governor; 
\item auxiliary verbs: an auxiliary verb may amend mood, tense and voice properties (active vs. passive construction);
\item coordination chains: the presence of a coordinate conjunct may create new morphological properties to the regent node, taking into account the properties of all involved nodes in the coordination chains. These new properties are called \textit{tree-gender}, \textit{tree-number} and \textit{tree-person} and may be used in order to establish some required future agreement (e.g. subject-predicate agreement).\footnote{The constraint may be fine-tuned and handle special case of \textit{appositive coordination} that does not implicate plural number.}
\end{itemize}

In the new hybrid approach, the features transportation mechanism doesn't require any approximation of high-order constraints (McCrae et al., 2008), which are instead indispensable in the original WCDG approach. The reason for this is that in wcdg-parsing the heuristic algorithm need to be informed of the structural changes (including node properties resulted by a transport) in order to do proper transformations in case of conflicts, and the complexity grows polynomially in the number of considered nodes.\footnote{In WCDG, the restriction to binary constraints (i.e. dependency relations between two nodes), motivated by computational issue, is a severe limitation of the expressiveness of the formalism.} In the new approach the analysis process is mainly data-driven exploiting the statistical language model, and the rich information resulting of \textit{constraints application} can be used at least for two different purposes:

\begin{enumerate}
\item by the search algorithm, as a ``grip'' to recover wrong transitions, for example through the optimistic back tracking technique (Ytrestøl, 2011);
\item as an additional features for the statistical component, for example to inform the parser that two \textit{focus} words are particularly ``connected'', also predicting early a possible dependency label, or that contrariwise they don't share any syntactic and semantic relation.
\end{enumerate}

For what concerns the features used in the statistical component, I believe that in order to obtain a robust parser they should be almost entirely delexicalized, except for the functional words where the lemma can be viewed  as a fine-grained part-of-speech information.\footnote{Unlexicalized parsing is also considered to be robust for applications such as cross-lingual parsing (McDonald et al., 2011)} This is necessary also for the number of lexical ambiguities that does not easily allow a discrete features representation. The new parser performs together, at the same time, syntactic and morphological analysis, so during the parsing process the words not yet fully analyzed keep all their possible lexical readings alive, creating, for instance, several features with different pos-tags candidates for each word.

Therefore, instead of the use of sparse \textit{forms} or {lemmas} features, which represent ``semantic instance'', in the new parser I found more appropriate to use some score of \textit{semantic relations} obtained exploiting a large \textit{vectorial semantic space} that records syntactic dependencies (e.g. subj-verb,  dobj-verb, pp-attachment). This structured knowledge could be generated using the \textit{learning by reading} technique, where a basic parser reads a large quantities of text in order to create information for the second parser.\footnote{To encode syntactic dependencies, instead of words order, I used the technique described in Basile (2011).}  Using also a stable semantic knowledge, it is possible improve syntactic parsing following some specific insights of Christen (2013).

Thanks to the active use of the constraints, the new parser is able to insert traces in a bottom-up way, in particular for missing arguments (e.g. for modal verbs). The mechanism of traces has been extended to cover more complex phenomena, such as gaps (e.g. where a verb is missing as during a coordination). After the traces integrations, the parser uses the same \textit{vector semantic space} to resolve some anaphoric references.

\section{Conclusion}

In this paper I explained the reasons that led me to research and conceive a novel a novel technology for dependency parsing. In particular I highlight the problem to infer the reliability of the results of a data-driven transition-based parser, which is extremely important for high-level processes that expect to use correct parsing results.

In my research I deeply investigated the use of different kinds of ``knowledge'' (e.g. syntactic constraints, rich morphology, punctuation, null elements, semantics, linguistic universals, temporal and spatial dimensions) in the same syntactic analysis, and identified a number of areas of possible actions. I then briefly introduced a number of notes about a new hybrid approach for dependency parsing that combines data-driven transition-based and constraint-based approaches. The result is a parser that proceeds with the analysis in a ``more aware'' way, that attempts to understand when it fails to analyze, maintaining a robust behavior and high efficiency. The aim is that a parser accepts not only ``well-formed' sentences but also deviant structures if no other analysis is feasible.

I realised that a perfect interoperability between software and data (i.e. treebank, dictionaries and rules) is crucial\footnote{Let me give you an analogy: it is known that Apple's superiority is due to its perfect hardware and software integration.}. I have encoded the Italian lexicon (subset of) in a formalism inspired by the Slot Grammar (SG) of McCord (1980) wherein every lexical entry contains information about category, morphological features and a set of slots and rules for filling them (e.g. words order, agreement information). This lexicon includes subcategorization for nouns, adjectives, verbs and adverbs. A great effort has been required to define the constraints grammar. In my research activity I could take advantage of Lisp implementation of a particular rule-based parser developed by Lesmo (2009), that got the best result (LAS 88.73\%) at Evalita 2009 DPT, and Lector developed by Christen (1990), from which I extracted some rules that I then transformed in a set of WCDG constraints.\footnote{See Christen (2013) for more information about a new version of Lector, called Syntagma, which implements linguistic constraints in a way similar to Property Grammar (Blache, 2006).} All the constraints have been then fine-tuned on the TUT Treebank and were tested against Evalita 2011 DPT, obtaining an attachment score of 96.16\%, the best result so far for a dependency parser for the Italian language.

Because of the intimate relationship between pos-tag and dependency label used in the new parser, I developed a special treebank that consists in 6,515 Italian sentences and 108,973 words, aligned with the syntactic relations described into the constraints and with the lexical readings resulting from the morphological analyzer. This treebank and the constraints, as well as the criteria for establishing dependency relations, are developed with the Tesni\`ere's four \textit{fundamental categories} (1959) in mind, in which only semantical \textit{full words} are allowed to behave as governor, that is: \textit{verbs}, \textit{nouns}, \textit{adjectives}, \textit{adverbs}. Therefore, the parser output it is closer to the \textit{collapsed} version of Stanford dependencies, where, for example, a \textit{prep} can't be a governor, so it is natively more suitable for Information Extraction. 

Some resources I developed during these research activities, including the Italian Treebank, are freely available on my GitHub repository.\footnote{https://github.com/matteo-grella/mg-research}

I like to believe that these few notes of mine could be a starting point for new researches in the field of dependency parsing.
 
\section*{Acknowledgments}

This research in its early stages was partially supported by Universit\`a degli Studi di Torino thanks to Leonardo Lesmo to whom I am particularly grateful. A special thanks also to Daniel Christen who have provided insights and expertise to this research from the beginning, and to Giuseppe Attardi who were a great sources of inspiration. I am also immensely grateful to Marco Nicola who provided invaluable help in transforming several ideas briefly sketched out into a system that actually works. I also thank Chiara Schilir\`o who generously helped me with the annotation of necessary linguistic resources to experiment with the ideas behind this research. Francesco Sabatini and Vittorio Coletti for some explanation about Valency Grammar. Last but not the least, Francesco Sartorio with whom I recently had a lot of never-ending always productive conversations about the most recent syntactic parsing algorithms and related machine learning systems.

\bibliographystyle{acl}
\nocite{*}
\bibliography{eacl2014}

\end{document}